\documentclass[11pt,a4paper]{article}
\usepackage[hyperref]{acl2019}
\usepackage{acl2019}
\usepackage{times}
\usepackage{latexsym}
\usepackage{url}
\usepackage{multirow}
\usepackage{adjustbox}
\usepackage[ruled]{algorithm2e} 
\usepackage{amssymb}
\usepackage{pifont}
\usepackage{url}
\usepackage{mwe}
\usepackage{amsmath}
\usepackage{amssymb}
\usepackage{wasysym}
\usepackage{bbm}
\usepackage{todonotes} 
\usepackage{graphicx}
\usepackage{caption}
\usepackage[caption=false]{subfig}

\aclfinalcopy 

\DeclareGraphicsExtensions{.pdf,.png,.jpg}
\newcommand\MNAME{MT-DNN}
\newcommand\NMNAME{MT-DNN\textsubscript{KD}}
\newcommand\SLABEL{target}

\title{Improving Multi-Task Deep Neural Networks via Knowledge Distillation for Natural Language Understanding}

\author{Xiaodong Liu$^1$, Pengcheng He$^2$, Weizhu Chen$^2$, Jianfeng Gao$^1$ \\
  $^1$ Microsoft Research~~~~~~~~
  $^2$ Microsoft Dynamics 365 AI \\
  {\tt \{xiaodl,penhe,wzchen,jfgao\}@microsoft.com}
}
\date{}

\begin{document}
\maketitle

\begin{abstract}

This paper explores the use of knowledge distillation to improve a Multi-Task Deep Neural Network (MT-DNN) \cite{liu2019mt-dnn} for learning text representations across multiple natural language understanding tasks. Although ensemble learning can improve  model performance, serving an ensemble of large DNNs such as MT-DNN can be prohibitively expensive. Here we apply the knowledge distillation method \cite{hinton2015distilling} in the multi-task learning setting. 
For each task, we train an ensemble of different MT-DNNs (teacher) that outperforms any single model, 
and then train a single MT-DNN (student) via multi-task learning to \emph{distill} knowledge from these ensemble teachers. 
We show that the distilled MT-DNN significantly outperforms the original MT-DNN on 7 out of 9 GLUE tasks, pushing the GLUE benchmark (single model) to 83.7\% (1.5\% absolute improvement\footnote{ Based on the GLUE leaderboard at https://gluebenchmark.com/leaderboard as of April 1, 2019.}). 
The code and pre-trained models will be made publicly available at https://github.com/namisan/mt-dnn. 
\end{abstract}

\section{Introduction}
\label{sec:introduction}

Ensemble learning is an effective approach to improve model generalization, and has been used to achieve new state-of-the-art results in a wide range of natural language understanding (NLU) tasks, including question answering and machine reading comprehension \cite{bert2018, liu2018san, huang2017fusionnet,hancock2019snorkel}. A recent survey is included in \cite{gao2019neural}. 
However, these ensemble models typically consist of tens or hundreds of different deep neural network (DNN) models and are prohibitively expensive to deploy due to the computational cost of runtime inference.
Recently, large-scale pre-trained models, such as BERT \cite{bert2018} and GPT \cite{gpt2018}, have been used effectively as the base models for building task-specific NLU models via fine-tuning.
The pre-trained models by themselves are already expensive to serve at runtime (e.g. BERT contains 24 transformer layers with 344 million parameters, and GPT-2 contains 48 transformer layers with 1.5 billion parameters), the ensemble versions of these models multiplying the extreme for online deployment. 

Knowledge distillation is a process of distilling or transferring the knowledge from a (set of) large, cumbersome model(s) to a lighter, easier-to-deploy single model, without significant loss in performance  \cite{buciluǎ2006model,hinton2015distilling,balan2015bayesian,ba2016layer,chen2015net2net,tan2018multilingual}.

In this paper, we explore the use of knowledge distillation to improve a Multi-Task Deep Neural Network (MT-DNN) \cite{liu2019mt-dnn} for learning text representations across multiple NLU tasks. 
Since MT-DNN incorporates a pre-trained BERT model, its ensemble is expensive to serve at runtime.

We extend the knowledge distillation method \cite{hinton2015distilling} to the multi-task learning setting \cite{caruana1997multitask,xu2018mtl, collobert2011natural, zhang2017survey,liu2015mtl}.
In the training process, we first pick a few tasks, each with an available task-specific training dataset which is stored in the form of $(x, y)$ pairs, where $x$ is an input and $y$ is its correct {\SLABEL}.
For each task, we train an ensemble of MT-DNN models (teacher) that outperform the best single model. Although the ensemble model is not feasible for online deployment, it can be utilized, in an offline manner, to produce a set of \emph{soft {\SLABEL}s} for each $x$ in the training dataset , which, for example, in a classification task are the class probabilities averaged over the ensemble of different models. 
Then, we train a single MT-DNN (student) via multi-task learning with the help of the teachers by using both the soft {\SLABEL}s and correct {\SLABEL}s across different tasks. We show in our experiments that knowledge distillation effectively transfers the generalization ability of the teachers to the student. As a result, the distilled MT-DNN outperforms the vanilla MT-DNN that is trained in a normal way, as described in \cite{liu2019mt-dnn}, on the same training data as was used to train the teachers.


We validate the effectiveness of our approach on the General Language Understanding Evaluation (GLUE) dataset \cite{wang2018glue} which consists of 9 NLU tasks. We find that the distilled MT-DNN outperforms the vanilla MT-DNN on 7 tasks, including the tasks where we do not have teachers.  This distilled model improves the GLUE benchmark (single model) to 83.7\%, amounting to 3.2\% absolute improvement over BERT and 1.5\% absolute improvement over the previous state of the art model based on the GLUE leaderboard\footnote{https://gluebenchmark.com} as of April 1, 2019.

In the rest of the paper, Section 2 describes the MT-DNN of \citet{liu2019mt-dnn} which is the baseline and vanilla model for this study. Section 3 describes in detail knowledge distillation for multi-task learning. Section 4 presents our experiments on GLUE. Section 5 concludes the paper.

\section{MT-DNN}
\label{sec:mt-dnn}

\begin{figure*}
	\centering
	\vspace{-1mm}
	 \adjustbox{trim={0.0\width} {0.01\height} {0.\width} {0.\height},clip}
     {
 	\includegraphics[width=0.92\textwidth]{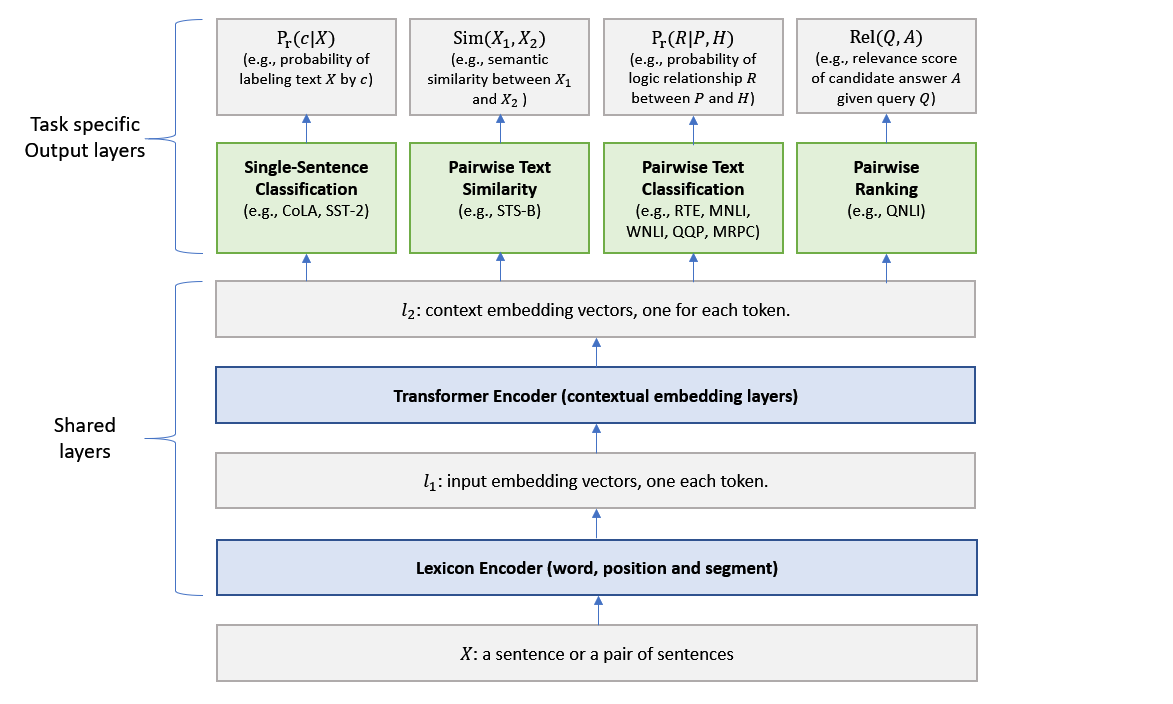}
     }
	\caption{Architecture of the MT-DNN model for representation learning \cite{liu2019mt-dnn}. The lower layers are shared across all tasks while the top layers are task-specific. The input $X$ (either a sentence or a set of sentences) is first represented as a sequence of embedding vectors, one for each word, in $l_1$. Then the Transformer encoder captures the contextual information for each word and generates the shared contextual embedding vectors in $l_2$. Finally, for each task, additional task-specific layers generate task-specific representations, followed by operations necessary for classification, similarity scoring, or relevance ranking.}
	\label{fig:mt-dnn}
\end{figure*}

The architecture of the MT-DNN model is shown in Figure \ref{fig:mt-dnn}. The lower layers are shared across all tasks, while the top layers represent task-specific outputs. The input $X$, which is a word sequence (either a sentence or a set of sentences packed together) is first represented as a sequence of embedding vectors, one for each word, in $l_1$. Then the transformer encoder captures the contextual information for each word via self-attention, and generates a sequence of contextual embeddings in $l_2$. This is the shared semantic representation that is trained by our multi-task objectives.  

\paragraph{Lexicon Encoder ($l_1$):} 
The input $X=\{x_1,...,x_m\}$ is a sequence of tokens of length $m$. Following \citet{bert2018}, the first token $x_1$ is always the \texttt{[CLS]} token. 
If $X$ is packed by a set of sentences $(X_1, X_2)$, we separate the these sentences with special tokens \texttt{[SEP]}. The lexicon encoder maps $X$ into a sequence of input embedding vectors, one for each token, constructed by summing the corresponding word, segment, and positional embeddings.

\paragraph{Transformer Encoder ($l_2$):}
We use a multi-layer bidirectional Transformer encoder \citep{vaswani2017attention} to map the input representation vectors ($l_1$) into a sequence of contextual embedding vectors 
$\mathbf{C} \in \mathbb{R}^{d \times m}$. 
This is the shared representation across different tasks. 

\paragraph{Task-Specific Output Layers:} 
We can incorporate arbitrary natural language tasks, each with its task-specific output layers. For example, we implement the output layers as a neural decoder for text generation, a neural ranker for relevance ranking, a logistic regression for text classification, and so on. Below, we elaborate the implementation detail using text classification as an example.  

Suppose that $\mathbf{x}$ is the contextual embedding ($l_2$) of the token \texttt{[CLS]}, which can be viewed as the semantic representation of input sentence $X$. The probability that $X$ is labeled as class $c$ (i.e., the sentiment is postive or negative) is predicted by a logistic regression with softmax:
\begin{equation}
P_r(c|X)= \text{softmax} (\mathbf{W}_{t} \cdot \mathbf{x}),
\label{eqn:single-sent-classification}
\end{equation}
where $\mathbf{W}_{t}$ is the task-specific parameter matrix for task $t$.

\subsection{The Training Procedure}

The training procedure of MT-DNN consists of two stages: pre-training and multi-task learning (MTL). In the pre-training stage, \citet{liu2019mt-dnn} used a publicly available pre-trained BERT model to initialize the parameters of the shared layers (i.e., the lexicon encoder and the transformer encoder).

In the MTL stage, mini-batch based stochastic gradient descent (SGD) is used to learn the model parameters (i.e., the parameters of all the shared layers and the task-specific layers), as shown in Algorithm \ref{algo:mtdnn}. First, the training samples from multiple tasks (e.g., 9 GLUE tasks) are packed into mini-batches. We denote a mini-batch by $b_t$, indicating that it contains only the samples from task $t$. In each epoch, a mini-batch $b_t$ is selected, and the model is updated according to the task-specific objective for task $t$, denoted by $L_t(\Theta)$. This approximately optimizes the sum of all multi-task objectives. 

Take text classification as an example. We use the cross-entropy loss as the objective in Line 3 of Algorithm \ref{algo:mtdnn}:
\begin{equation}
-\sum_c \mathbbm{1}(X,c) \log(P_r(c|X)),
\label{eqn:cross-entropy-loss}
\end{equation}
where $\mathbbm{1}(X,c)$ is the binary indicator (0 or 1) if class label $c$ is the correct classification for $X$, and $P_r(.)$ is defined by Equation \ref{eqn:single-sent-classification}.

Then, in Line 5, the parameters of the shared layers and the output layers corresponding to task $t$ are updated using the gradient computed in Line 4.

After MT-DNN is trained via MTL, it can be fine-tuned (or adapted) using task-specific labeled training data to perform prediction on any individual task, which can be a task used in the MTL stage or a new task that is related to the ones used in MTL. 
\citet{liu2019mt-dnn} showed that the shared layers of MT-DNN produce more universal text representations than that of BERT. As a result, MT-DNN allows fine-tuning or adaptation with substantially fewer task-specific labels.

\begin{algorithm}[ht!]
 \SetAlgoLined
Initialize model parameters $\Theta$ randomly.  \\
Initialize the shared layers (i.e., the lexicon encoder and the transformer encoder) using a pre-trained BERT model. \\
Set the max number of epoch: $epoch_{max}$.
\textit{//Prepare the data for $T$ tasks.}\\
\For{$t$ in $1,2,...,T$ }
{
    Pack the dataset $t$ into mini-batch: $D_t$.
}

 \For{$epoch$ in $1,2,...,epoch_{max}$}{
     1. Merge all the datasets: $D =D_1 \cup D_2 ... \cup D_T$ \\
     2. Shuffle $D$ \\
     \For{$b_t$ in D}{
        \textit{//$b_t$ is a mini-batch of task $t$.} \\
     3. Compute task-specific loss : $L_t(\Theta)$ \\
     4. Compute gradient: $\nabla(\Theta)$ \\
     5. Update model: $\Theta = \Theta - \epsilon \nabla(\Theta)$ \\
     }
 }
 \caption{\label{algo:mtdnn} Training a MT-DNN model.}
\end{algorithm} \vspace{-0.2cm}

\section{Knowledge Distillation}
\label{sec:knowledge-distillation}

\begin{figure*}
	\centering
	 \adjustbox{trim={0.0\width} {0.01\height} {0.\width} {0.\height},clip}
     {
 	\includegraphics[width=0.66\textwidth]{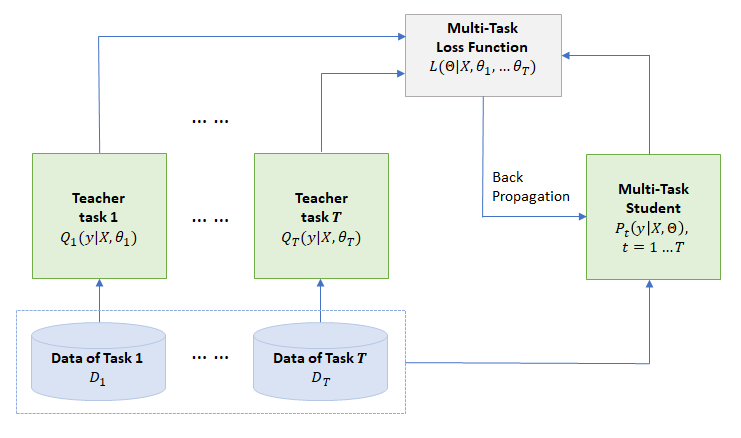}
     }
	\caption{Process of knowledge distillation for multi-task learning. A set of tasks where there is task-specific labeled training data are picked. Then, for each task, an ensemble of different neural nets (teacher) is trained. The teacher is used to generate for each task-specific training sample a set of soft targets.  Given the soft targets of the training datasets across multiple tasks, a single MT-DNN (student) is trained using multi-task learning and back propagation as described in Algorithm \ref{algo:mtdnn}, except that if task $t$ has a teacher, the task-specific loss in Line 3 is the average of two objective functions, one for the correct targets and the other for the soft targets assigned by the teacher.
}
	\label{fig:knowledge-distillation}
\end{figure*}

The process of knowledge distillation for MTL is illustrated in Figure \ref{fig:knowledge-distillation}.  
First, we pick a few tasks where there are task-specific labeled training data. 
Then, for each task, we train an ensemble of different neural nets as a teacher. Each neural net is an instance of MT-DNN described in Section \ref{sec:mt-dnn}, and is fine-tuned using task-specific training data while the parameters of its shared layers are initialized using the MT-DNN model pre-trained on the GLUE dataset via MTL, as in Algorithm \ref{algo:mtdnn}, and the parameters of its task-specific output layers are randomly initialized. 

For each task, a teacher generates a set of soft targets for each task-specific training sample. 
Take text classification as an example. 
A neural network model typically produces class probabilities using a softmax layer as in Equation \ref{eqn:single-sent-classification}. Let $Q^k$ be the class probabilities produced by the $k$-th single network of the ensemble. The teacher produces the soft targets by averaging the class probabilities across networks:
\begin{equation}
Q = \text{avg} ([Q^1, Q^2, ..., Q^{K}]).
\label{eqn:teacher}
\end{equation}

We want to approximate the teacher using a student neural network model, which also has a softmax output for the same task $P_r(c|X)$, as in Equation \ref{eqn:single-sent-classification}. Hence, we use the standard cross entropy loss:
\begin{equation}
-\sum_c Q(c|X) \log(P_r(c|X)).
\label{eqn:cross-entropy-loss-soft}
\end{equation}

Note that the above loss function differs from the cross entropy loss in Equation \ref{eqn:cross-entropy-loss} in that the former uses the soft targets $Q(c|X)$ while the latter uses the hard correct target via the indicator $\mathbbm{1}(X,c)$. 

As pointed out by \citet{hinton2015distilling}, the use of soft targets produced by the teacher is the key to successfully transferring the generalization ability of the teacher to the student. The relative probabilities of the teacher labels contain information  
about how the teacher generalizes. For example, the sentiment of the sentence ``I really enjoyed the conversation with Tom'' has a small chance of being classified as negative. 
But the sentence ``Tom and I had an interesting conversation'' can be either positive or negative, depending on its context if available, leading to a high entropy of the soft targets assigned by the teacher. 
In these cases, the soft targets provide  more information per training sample than the hard targets and less variance in the gradient between training samples. 
By optimizing the student for the soft targets produced by the teacher, we expect the student to learn to generalize in the same way as the teacher.  In our case, each task-specific teacher is the average of a set of different neural networks, and thus generalizes well. The single MT-DNN (student) trained to generalize in the same way as the teachers is expected to do much better on test data than the vanilla MT-DNN that is trained in the normal way on the same training dataset. We will demonstrate in our experiments that this is indeed the case. 

We also find that when the correct targets are known, the model performance can be significantly improved by training the distilled model on a combination of \textit{soft} and \textit{hard} targets. 
We do so by defining a loss function for each task  that take a weighted average between the cross entropy loss with the correct targets as Equation \ref{eqn:cross-entropy-loss} and the cross entropy with the soft targets as Equation \ref{eqn:cross-entropy-loss-soft}. 
\citet{hinton2015distilling} suggested using a considerably lower weight on the first loss term. 
But in our experiments we do not observe any significant difference by using different weights for the two loss terms, respectively.

Finally, given the soft targets of the training datasets across multiple tasks, the student MT-DNN can be trained using MTL as described in Algorithm \ref{algo:mtdnn}, except that if task $t$ has a teacher, the task-specific loss in Line 3 is the average of two objective functions, one for the correct targets and the other for the soft targets assigned by the teacher.

\section{Experiments}
\label{sec:exp}

\begin{table*}[htb!]
	\begin{center}
		\begin{tabular}{l|l|c|c|c|c|c}
			\hline \bf Corpus &Task& \#Train & \#Dev & \#Test   & \#Label &Metrics\\ \hline \hline
			\multicolumn{6}{@{\hskip1pt}r@{\hskip1pt}}{Single-Sentence Classification (GLUE)} \\ \hline
			CoLA & Acceptability&8.5k & 1k & 1k & 2 & Matthews corr\\ \hline
			SST-2 & Sentiment&67k & 872 & 1.8k & 2 & Accuracy\\ \hline \hline
			\multicolumn{6}{@{\hskip1pt}r@{\hskip1pt}}{Pairwise Text Classification (GLUE)} \\ \hline
			MNLI & NLI& 393k& 20k & 20k& 3 & Accuracy\\ \hline
            RTE & NLI &2.5k & 276 & 3k & 2 & Accuracy \\ \hline
            WNLI & NLI &634& 71& 146& 2 & Accuracy \\ \hline
			QQP & Paraphrase&364k & 40k & 391k& 2 & Accuracy/F1\\ \hline
            MRPC & Paraphrase &3.7k & 408 & 1.7k& 2&Accuracy/F1\\ \hline
			\multicolumn{5}{@{\hskip1pt}r@{\hskip1pt}}{Text Similarity (GLUE)} \\ \hline
			STS-B & Similarity &7k &1.5k& 1.4k &1 & Pearson/Spearman corr\\ \hline

\multicolumn{6}{@{\hskip1pt}r@{\hskip1pt}}{Relevance Ranking (GLUE)} \\ \hline \hline
			QNLI & QA/NLI& 108k &5.7k&5.7k&2& Accuracy\\ \hline \hline

		\end{tabular}
	\end{center}
	\caption{Summary of the GLUE benchmark.
	}
	\label{tab:datasets}
\end{table*}

We evaluate the MT-DNN trained using Knowledge Distillation, termed as {\NMNAME} in this section, on the General Language Understanding Evaluation (GLUE) benchmark. GLUE is a collection of nine NLU tasks as in Table~\ref{tab:datasets}, including question answering, sentiment analysis, text similarity and textual entailment. 
We refer readers to \citet{wang2018glue} for a detailed description of GLUE.
We compare {\NMNAME} with existing state-of-the-art models including BERT \cite{bert2018}, STILT  \cite{phang2018sentence}, Snorkel MeTal \cite{hancock2019snorkel}, and {\MNAME} \cite{liu2019mt-dnn}. Furthermore, we investigate the relative contribution of using knowledge distillation for MTL with an ablation study.

\subsection{Implementation details}
\label{subsec:impl}
Our implementation is based on the PyTorch implementations of  MT-DNN\footnote{https://github.com/namisan/mt-dnn} and BERT\footnote{https://github.com/huggingface/pytorch-pretrained-BERT}.
We used Adamax \cite{kingma2014adam} as our optimizer with a learning rate of 5e-5 and a batch size of 32. The maximum number of epochs was set to 5. 
A linear learning rate decay schedule with warm-up over 0.1 was used, unless stated otherwise.
We also set the dropout rate of all the task-specific layers as 0.1, except 0.3 for MNLI and 0.05 for CoLA/SST-2. 
To avoid the gradient explosion issue, we clipped the gradient norm within 1. 
All the texts were tokenized using wordpieces, and were chopped to spans no longer than 512 tokens.

To obtain a set of diverse single models to form ensemble models (teachers), we first trained 6 single {\MNAME}s, initialized using Cased/Uncased BERT models as \cite{hancock2019snorkel} with a different dropout rate, ranged in \{0.1, 0.2, 0.3\}, on the shared layers, while keeping other training hyperparameters the same as aforementioned. 
Then, we selected top 3 best models according to the results on the MNLI and RTE development datasets. 
Finally, we fine-tuned the 3 models on each of the MNLI, QQP, RTE and QNLI tasks to form four task-specific ensembles (teachers), each consisting of 3 single MT-DNNs fine-tuned for the task.
The teachers are used to generate soft {\SLABEL}s for the four tasks as Equation \ref{eqn:teacher}, described in Section \ref{sec:knowledge-distillation}. 
We only pick four out of nine GLUE tasks to train teachers  to investigate the generalization ability of {\NMNAME}, i.e., its performance on the tasks with and without teachers.

\begin{table*}[htb!]
\small
	\begin{center}
		\begin{tabular}{l|@{\hskip1pt}l@{\hskip1pt}|@{\hskip1pt}c@{\hskip1pt}|@{\hskip1pt}c@{\hskip1pt}|@{\hskip1pt}c@{\hskip1pt}|@{\hskip1pt}c@{\hskip1pt}|@{\hskip1pt}c|@{\hskip1pt}c|@{\hskip1pt}c |@{\hskip1pt} c |@{\hskip1pt} c|@{\hskip1pt} c}
			\hline \bf Model &CoLA&	SST-2 &MRPC& STS-B&QQP&MNLI-m/mm&QNLI&RTE&WNLI&AX &\textbf{Score}\\ 
			& 8.5k &67k &3.7k &7k &364k &393k &108k &2.5k &634 & & \\ \hline \hline
			BiLSTM+ELMo+Attn $^1$&36.0 &90.4 &84.9/77.9 &75.1/73.3 &64.8/84.7 &76.4/76.1 &79.8 &56.8 &65.1 &26.5 &70.0 \\ \hline
			\begin{tabular}{@{}c@{}}Singletask Pretrain \\Transformer $^2$   \end{tabular}
			 &45.4 &91.3 &82.3/75.7&82.0/80.0 &70.3/88.5 &82.1/81.4 &87.4 &56.0 &53.4  &29.8 &72.8 \\ \hline
			GPT on STILTs $^3$ &47.2 &93.1 &87.7/83.7 &85.3/84.8 &70.1/88.1 &80.8/80.6 &- &69.1 &65.1 &29.4 &76.9 \\ \hline
			BERT$_{\text{LARGE}}$$^4$ & 60.5 &94.9 &89.3/85.4 &87.6/86.5 &72.1/89.3 &86.7/85.9 &92.7 &70.1 &65.1	&39.6 & 80.5\\ \hline
        {\MNAME}$^5$ & 61.5 &95.6 &90.0/86.7 &88.3/87.7 &72.4/89.6 &86.7/86.0	&- &75.5 &65.1	&40.3 &82.2 \\ \hline
Snorkel MeTaL $^6$ &63.8&	\textbf{96.2}&	91.5/88.5&	\textbf{90.1/89.7}&	73.1/89.9&	87.6/87.2&93.9&80.9&65.1&39.9 &83.2 \\ \hline	        
        ALICE $^*$& 63.5&	95.2&	\textbf{91.8/89.0}&	89.8/88.8&	 \textbf{74.0/90.4}&	\textbf{87.9/87.4}&	95.7&	80.9&	65.1&	40.7& 83.3 \\ \hline
			\textbf{\NMNAME} &\textbf{65.4}   &95.6  &91.1/88.2 &89.6/89.0 &72.7/89.6 &87.5/86.7 & \textbf{96.0} & \textbf{85.1}&65.1 &\textbf{42.8} &\textbf{83.7} \\ \hline \hline
			{Human Performance} &66.4&97.8&86.3/80.8    &92.7/92.6	&59.5/80.4	&92.0/92.8	&91.2	&93.6	&95.9 &- & 87.1\\ \hline
		\end{tabular}
	\end{center}
	\caption{GLUE test set results scored using the GLUE evaluation server. The number below each task denotes the number of training examples. The state-of-the-art results are in \textbf{bold}. {\NMNAME} uses BERT\textsubscript{LARGE} to initialize its shared layers. 
	All the results are obtained from \href{https://gluebenchmark.com/leaderboard}{https://gluebenchmark.com/leaderboard} on April 1, 2019. Note that Snorkel MeTaL is an ensemble model.  
	- denotes the missed result of the latest GLUE version. 
	$^*$ denotes the unpublished work, thus not knowing whether it is a single model or an ensemble model. 
	For QNLI, we treat it as two tasks, pair-wise ranking and classification task on v1 and v2 training datasets, respectively, and then merge results on the test set.  
	Model references: $^1$:\protect\cite{wang2018glue} ; $^2$:\protect\cite{gpt2018}; $^3$: \protect\cite{phang2018sentence}; $^4$:\protect\cite{bert2018}; $^5$: \protect\cite{liu2019mt-dnn}; $^6$: \protect\cite{hancock2019snorkel}. 
	}
	\label{tab:glue_test}
\end{table*}

\begin{table*}
	\begin{center}
		\begin{tabular}{@{\hskip1pt}l|c@{\hskip1pt}|c@{\hskip1pt}|c@{\hskip1pt}|c@{\hskip1pt}|c@{\hskip1pt}|@{\hskip1pt}c|c @{\hskip1pt}|c@{\hskip1pt}}
			\hline \bf Model &MNLI-{m/mm} &QQP &RTE &QNLI(v2) &MRPC &CoLa &SST-2  &STS-B\\ \hline \hline
			BERT$_{\text{LARGE}}$& 86.3/86.2 &91.1/88.0 &71.1 &92.4 &89.5/85.8 &61.8 &93.5 &89.6/89.3\\
			\hline
            {\MNAME}  &87.1/86.7 &\textbf{91.9}/89.2 &83.4 &92.9 &91.0/87.5 &63.5 &\textbf{94.3} &90.7/90.6\\ \hline
            {\NMNAME} &\textbf{87.3/87.3} & \textbf{91.9/89.4} & \textbf{88.6}& \textbf{93.2} &\textbf{93.3/90.7} &\textbf{64.5} & \textbf{94.3}& \textbf{91.0/90.8}\\ \hline \hline
            {\MNAME-ensemble}  &88.1/87.9 &92.5/90.1 &86.7 &93.5 &\color{blue}{\textit{93.4/91.0}} &\color{blue}{\textit{64.5}} & \color{blue}{\textit{94.7}} &\color{blue}{\textit{92.1/91.6}}\\ \hline

		\end{tabular}
	\end{center}
	\caption{GLUE dev set results. The best result on each task produced by a single model is in \textbf{bold}.
	MT-DNN uses BERT\textsubscript{LARGE} as their initial shared layers. 
	{\NMNAME} is the MT-DNN trained using the proposed knowledge distillation based MTL. 
	MT-DNN-ensemble denotes the results of the ensemble models described in Section~\ref{subsec:impl}. The ensemble models on MNLI, QQP, RTE and QNLI are used as teachers in the knowledge distillation based MTL, while the other ensemble modes, whose results are in {\color{blue}{\textit{blue and italic}}}, 
	are not used as teachers.
	}
	\label{tab:glue_dev}
\end{table*}

\subsection{GLUE Main Results}
\label{subsec:results}
We compare {\NMNAME} with a list of state-of-the-art models that have been submitted to the GLUE leaderboard. 

\paragraph{BERT\textsubscript{LARGE}} This is the large BERT model released by \citet{bert2018}, which we used as a baseline. We used single-task fine-tuning to produce the best result for each GLUE task according to the development set. 

\paragraph{MT-DNN} This is the model described in Section \ref{sec:mt-dnn} and \citet{liu2019mt-dnn}. We used the pre-trained BERT\textsubscript{LARGE} model to initialize its shared layers, refined the shared layers via MTL on all GLUE tasks,
and then perform a fine-tune for each GLUE task using the task-specific data.
\paragraph{{\NMNAME}} This is the MT-DNN model trained using knowledge distillation as described in Section \ref{sec:knowledge-distillation}. 
{\NMNAME} uses the same model architecture as that of MT-DNN. But the former is trained with the help from four task-specific ensembles (teachers). {\NMNAME} is optimized for the multi-task objectives that are based on the hard correct {\SLABEL}s, as well as the soft {\SLABEL}s produced by the teachers if available. 
After knowledge distillation based MTL, {\NMNAME} is further fine-tuned for each task using task-specific data to produce the final predictions for each GLUE task on blind test data for evaluation.

The main results on the official test datasets of GLUE are reported in Table \ref{tab:glue_test}. 
Compared to other recent submissions on the GLUE leaderboard, {\NMNAME} is the best performer,
creating a new state-of-the-art result of 83.7\%. 
The margin between {\NMNAME} and the second-best model ALICE is 0.5\%,  larger than the margin of 0.1\% between the second and the third (and the fourth) places. 
It is worth noting that {\NMNAME} is a single model while Snorkel MetaL \cite{hancock2019snorkel} is an ensemble model. The description of ALICE is not disclosed yet.

Table \ref{tab:glue_test} also shows that {\NMNAME} significantly outperforms MT-DNN not only in overall score but on 7 out of 9 GLUE tasks, including the tasks without a teacher.
Since {\NMNAME} and MT-DNN use the same network architecture, and are trained with the same initialization and on the same datasets, the improvement of {\NMNAME} is solely attributed to the use of knowledge distillation in MTL. 

We note that the most significant per-task improvements are from CoLA (65.4\% vs. 61.5\%) and RTE (85.1\% vs. 75.5\%). Both tasks have relatively small amounts of in-domain data. Similarly, for the same type of tasks, the improvements of {\NMNAME} over {\MNAME} are much more substantial for the tasks with less in-domain training data e.g., for the two NLI tasks, the improvement in RTE is much larger than that in MNLI;  for the two paraphrase tasks, the improvement in  MRPC is larger than that in QQP. 
These results suggest that knowledge distillation based MTL is  effective at improving model performance for not only tasks with teachers but also ones without teachers, and more so for tasks with fewer in-domain labels.



\subsection{Ablation Study}

We perform an ablation study to investigate how effective it can distill knowledge from the ensemble models (teachers) to a single MT-DNN (student). To this end, we compare the performance of the ensemble models with the corresponding student model. 

The results on dev sets are shown in Table \ref{tab:glue_dev}, where MT-DNN-ensemble are the task-specific ensemble models trained using the process described in Section \ref{sec:knowledge-distillation}. We only use four ensemble models (i.e., the models for MNLI, QQP, RTE, QNLI) as teachers. The results of the other ensemble models (i.e., MRPC, CoLa, SST-2, STS-B) are reported to show the effectiveness of the knowledge distillation based MTL at improving the performance on tasks without a teacher.

We can draw several conclusions from the results in Table \ref{tab:glue_dev}. First, {\NMNAME} significantly outperforms MT-DNN and BERT\textsubscript{LARGE} across multiple GLUE tasks on the dev sets, which is consistent with what we observe on test sets in Table \ref{tab:glue_test}. 
Second, comparing {\NMNAME} with MT-DNN-ensemble, we see that the {\NMNAME} successfully distills knowledge from the teachers. Although the distilled model is  simpler than the teachers, it retains nearly all of the improvement that is achieved by the ensemble models. More interestingly, we find that incorporating knowledge distillation into MTL improves the model performance on the tasks where no teacher is used. On MRPC, CoLA, and STS-B, the performance of {\NMNAME} is much better than MT-DNN and is close to the ensemble models although the latter are not used as teachers in MTL.

\section{Conclusion}
\label{sec:con}

In this work, we have extended knowledge distillation to MTL in training a MT-DNN for natural language understanding.
We have shown that distillation works very well for transferring knowledge from a set of ensemble models (teachers) into a single, distilled MT-DNN (student). On the GLUE datasets, the distilled MT-DNN creates new state of the art result on 7 out of 9 NLU tasks, including the tasks where there is no teacher, pushing the GLUE benchmark (single model) to 83.7\%.

We show that the distilled MT-DNN retains nearly all of the improvements achieved by ensemble models, while keeping the model size the same as the vanilla MT-DNN model.


There are several research areas for future exploration. 
First, we will seek better ways of combining the soft targets and hard correct targets for multi-task learning. Second, the teachers might be used to produce the soft targets for large amounts of unlabeled data, which in turn can be used to train a better student model in a way conceptually similar to semi-supervised learning. 
Third, instead of compressing a complicated model to a simpler one, knowledge distillation can also be used to improve the model performance regardless of model complexity, in  machine learning scenarios such as self-learning in which both the student and teacher are the same model.





\section*{Acknowledgments}
We thank 
Asli Celikyilmaz,
Xuedong Huang,
Moontae Lee,
Chunyuan Li,
Xiujun Li,
and Michael Patterson
for helpful discussions and comments.


\bibliography{acl_snli}
\bibliographystyle{acl_natbib}
%

\end{document}


\onecolumn

\section{Appendices}
\label{sec:appendix}


\subsection{Test results on the \textit{old} GLUE test set}
\label{sec:glue_old}
\begin{table*}[h!]
\small
	\begin{center}
		\begin{tabular}{l|@{\hskip1pt}l@{\hskip1pt}|@{\hskip1pt}c@{\hskip1pt}|@{\hskip1pt}c@{\hskip1pt}|@{\hskip1pt}c@{\hskip1pt}|@{\hskip1pt}c@{\hskip1pt}|@{\hskip1pt}c|@{\hskip1pt}c|@{\hskip1pt}c |@{\hskip1pt} c |@{\hskip1pt} c|@{\hskip1pt} c}
			\hline \bf Model &CoLA&	SST-2 &MRPC& STS-B&QQP&MNLI-m/mm&QNLI&RTE&WNLI&AX &\textbf{Score}\\ 
			& 8.5k &67k &3.7k &7k &364k &393k &108k &2.5k &634 & & \\ \hline \hline
			BiLSTM+ELMo+Attn $^1$ &36.0 &90.4 &84.9/77.9 &75.1/73.3 &64.8/84.7 &76.4/76.1 &79.9 &56.8 &65.1 &26.5 &70.5 \\ \hline
			\begin{tabular}{@{}c@{}}Singletask Pretrain \\Transformer $^2$  \end{tabular}
			 &45.4 &91.3 &82.3/75.7&82.0/80.0 &70.3/88.5 &82.1/81.4 &88.1 &56.0 &53.4  &29.8 &72.8 \\ \hline
			GPT on STILTs $^3$ &47.2 &93.1 &87.7/83.7 &85.3/84.8 &70.1/88.1 &80.8/80.6 &87.2 &69.1 &65.1 &29.4 &76.9 \\ \hline
			BERT$_{\text{LARGE}}$ $^4$ & 60.5 &94.9 &89.3/85.4 &87.6/86.5 &72.1/89.3 &86.7/85.9 &91.1 &70.1 &65.1	&39.6 & 80.4\\ \hline
MT-DNN &\textbf{61.5} &\textbf{95.6} &\textbf{90.0/86.7} &\textbf{88.3/87.7} &\textbf{72.4/89.6} &\textbf{86.7/86.0}	&\textbf{98.0} &\textbf{75.5} &65.1	&\textbf{40.3} &\textbf{82.2} \\ \hline
		\end{tabular}
	\end{center}
	\caption{GLUE test set results, which are scored by the GLUE evaluation server. The numbers below each task denote the size of training examples. The state-of-the-art results are in \textbf{bold}. All the results are obtained from \href{https://gluebenchmark.com/leaderboard}{https://gluebenchmark.com/leaderboard} on January 15, 2019. Note that the \textit{old} version of GLUE test set expired on January 30, 2019. Model references: $^1$:\protect\cite{wang2018glue} ; $^2$:\protect\cite{gpt2018}; $^3$: \protect\cite{phang2018sentence};  $^4$:\protect\cite{bert2018}.
	}
	\label{tab:glue_test}
\end{table*}
\bibliography{acl_snli}
\bibliographystyle{acl_natbib}